\documentclass{article}
\usepackage{spconf,amsmath,graphicx}

\usepackage{enumitem}
\setlist{nosep, leftmargin=14pt}

\usepackage{mwe} 
\usepackage{cite}
\usepackage{amsmath,amssymb,amsfonts}
\usepackage{algorithmic}
\usepackage{graphicx}
\usepackage{textcomp}
\usepackage{lipsum}
\usepackage{float}
\usepackage{fontenc}
\usepackage{setspace}
\usepackage{multirow}
\usepackage{textcomp}
\usepackage{bm}
\usepackage{listings}
\usepackage[mathscr]{euscript}
\usepackage{multirow}
\usepackage{float}
\usepackage[table,xcdraw]{xcolor}
\usepackage{caption}
\usepackage{setspace}
\usepackage{booktabs}
\usepackage{hyperref}

\usepackage{bibspacing}

\begin{document}

\title{ICHPro: Intracerebral Hemorrhage Prognosis Classification via Joint-Attention Fusion-based 3D Cross-Modal Network}
%
\name{\begin{tabular}[t]{c}
Xinlei Yu$^1$, Xinyang Li$^2$, Ruiquan Ge$^{1,*}$, Shibin Wu$^3$, Ahmed Elazab$^4$, Jichao Zhu$^5$, Lingyan Zhang$^5$, \\Gangyong Jia$^1$, Taosheng Xu$^6$, Xiang Wan$^7$, Changmiao Wang$^{7,*}$
\thanks{$^{*}$Corresponding authors: Ruiquan Ge and Changmiao Wang.}
\end{tabular}}
\address{\fontsize{11pt}{5pt}\selectfont$^1$Hangzhou Dianzi University, China \quad
\fontsize{11pt}{5pt}\selectfont$^2$The Chinese University of Hong Kong, Shenzhen, China \\
\fontsize{11pt}{5pt}\selectfont$^3$Ping An Technology, China \quad
\fontsize{11pt}{5pt}\selectfont$^4$Shenzhen University, China \quad
\fontsize{11pt}{5pt}\selectfont$^5$Longgang Central Hospital of Shenzhen, China \\
\fontsize{11pt}{5pt}\selectfont$^6$Hefei Institutes of Physical Science, Chinese Academy of Sciences, China\\
\fontsize{11pt}{5pt}\selectfont$^7$Shenzhen Research Institute of Big Data, China}
%
%
%
%

\maketitle

\begin{abstract}
    Intracerebral Hemorrhage (ICH) is the deadliest subtype of stroke, necessitating timely and accurate prognostic evaluation to reduce mortality and disability. However, the multifactorial nature and complexity of ICH make methods based solely on computed tomography (CT) image features inadequate. Despite the capacity of cross-modal networks to fuse additional information, the effective combination of different modal features remains a significant challenge. In this study, we propose a joint-attention fusion-based 3D cross-modal network termed ICHPro that simulates the ICH prognosis interpretation process utilized by neurosurgeons. ICHPro includes a joint-attention fusion module to fuse features from CT images with demographic and clinical textual data. We introduce a joint loss function to enhance the representation of cross-modal features. ICHPro facilitates the extraction of richer cross-modal features, thereby improving classification performance. Upon testing our method using a five-fold cross-validation, we achieved an accuracy of \textbf{89.11\%}, an F1 score of \textbf{0.8767}, and an AUC value of \textbf{0.9429}. These results outperform those obtained from other advanced methods based on the test dataset, thereby demonstrating the superior efficacy of ICHPro.
The code is available at our github \footnote{Our source code is at: \href{https://github.com/YU-deep/ICH_prognosis.git}{https://github.com/YU-deep/ICH\_prognosis.git}}. 
\end{abstract}


\begin{keywords}
Joint-attention mechanism, Cross-modal fusion, Demographic and clinical text, ICH prognosis
\end{keywords}

\vspace{-0.10cm}
\section{Introduction}
Intracerebral Hemorrhage (ICH) carries an extremely high mortality rate of more than 40\%, with only 20\% of survivors achieving functional independence \cite{broderick2021story}. Consequently, accurate prognosis prediction is of crucial importance for patients post-ICH in order to develop an appropriate treatment plan \cite{rosand2021preserving}. Experienced neurosurgeons predominantly rely on computed tomography (CT) scans, specifically the location, volume, and distinct texture features of the hemorrhage site, as the primary determinants for judgment. Secondary indicators include the patient's age, gender, and Glasgow Coma Scale (GCS) score \cite{teasdale2014glasgow} among others \cite{troiani2021prognostic}. This process, however, is contingent on manual predictions by neurosurgeons, a labor-intensive task that may affect accuracy due to variability in doctors' experience and subjective factors. To address these issues, early studies have employed machine learning techniques \cite{ramos2020predicting,nawabi2021imaging}, achieving certain levels of success, albeit with room for further improvement.



Despite the richer and more comprehensive information that can be obtained with cross-modal methods, their application in ICH prognosis remains limited and preliminary. Recently, there have been some advances, such as the fusion-based \cite{wang2021data} and deep learning (DL)-based methods \cite{perez2023deep} that directly concatenated extracted image with clinical features. Also, GCS-ICHNet \cite{shan2023gcs} improves performance by fusing images with domain knowledge using a self-attention mechanism. However, these methods lack an effective fusion mechanism, limiting the establishment of semantic connections and internal dependencies of features between modalities.

In response to these limitations, in this paper we propose a novel method boasting four key benefits: (1) The 3D structure provides more spatial texture features of hemorrhage locations. (2) The cross-modal structure incorporates more comprehensive demographic and clinical data, thereby enhancing the model's understanding of the task. (3) The joint-attention mechanism directs the network to adjust regions of attention, facilitating the acquisition of richer and more effective fusion features. (4) The Vision-Text Modality Fusion (VTMF) loss, specifically designed for the cross-modal network, promotes better feature representations across the two modalities.

\begin{figure*}[ht]
    \begin{center}
    \includegraphics[width=14.5cm,height=6.22cm]{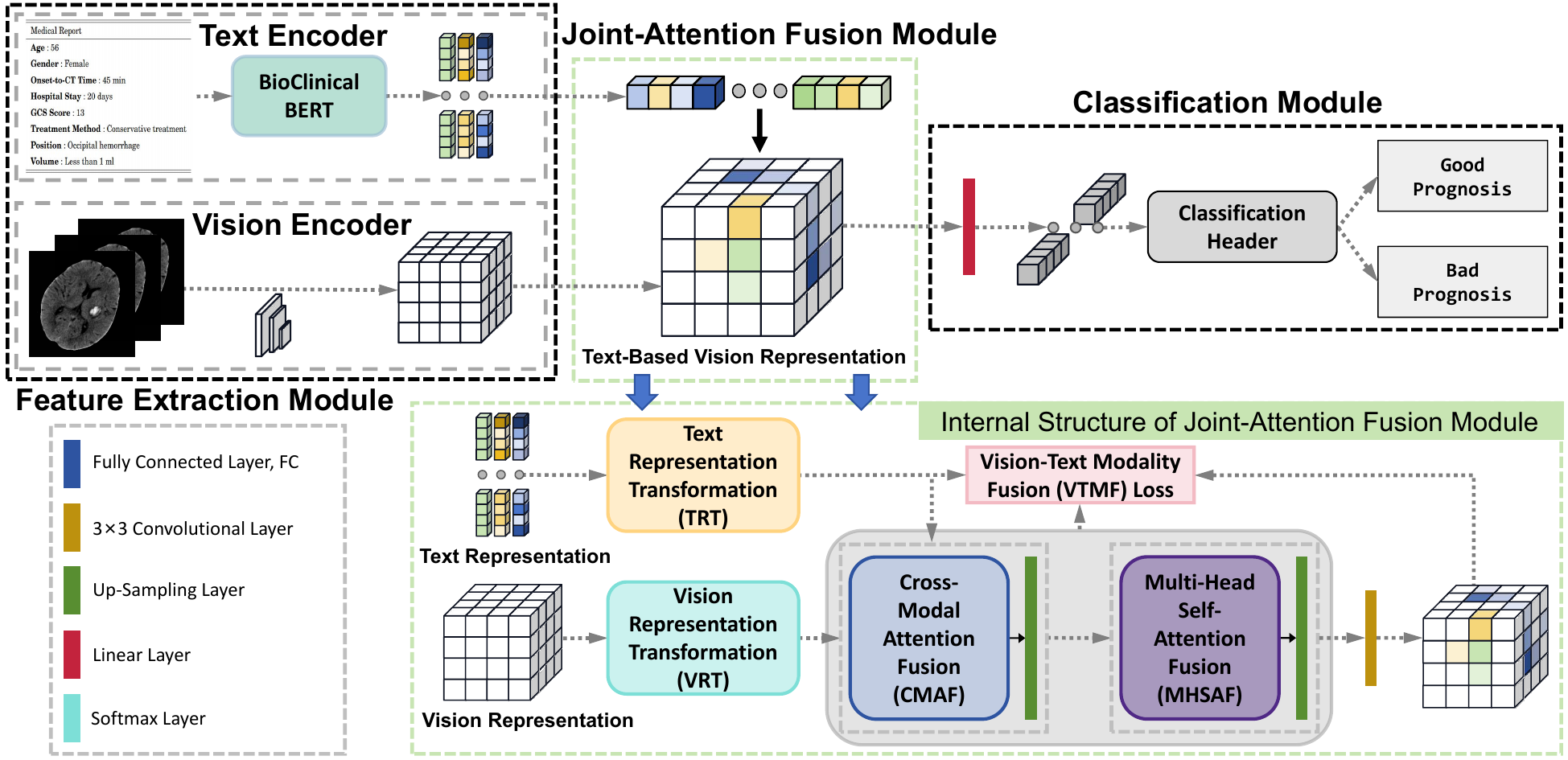}
    \end{center}
   \caption{The illustration delineated the architecture of ICHPro and the green-dashed box below represents the internal structure of the joint-attention fusion Module. The CMAF block is designed to facilitate the fusion of textual and visual modalities. Concurrently, the VTMF loss actively encourages the superior formation of representation of cross-modal features.}
    \label{overview}
\end{figure*}

\vspace{-0.20cm}
\section{METHODOLOGY}

As depicted in Fig.\ref{overview}, ICHPro comprises three components: the feature extraction module, the joint-attention fusion module, and the classification module. These modules represent the three consecutive stages of the entire process. 

In the feature extraction module, we employ the pre-trained BioClinicalBERT \cite{alsentzer-etal-2019-publicly} model as the text encoder to obtain text representation $f^t$, and the pre-trained 3D ResNet-50 \cite{hara2018can} as the vision encoder to secure vision representation $f^v$. In the classification module, the pre-trained 1D DenseNet-121 is utilized as the classification header.

\vspace{-0.15cm}
\subsection{Joint-Attention Fusion Module}
In this module, $f^t$ is first fed into the text representation transformation (TRT) block and $f^v$ into the vision representation transformation (VRT) block, respectively. This process yields a unified reconstructed text representation $\tilde{f^{t}}$ and reconstructed vision representation $\tilde{f^{v}}$. These are subsequently processed through a cross-modal attention fusion (CMAF) block and a multi-head self-attention fusion (MHSAF) block, respectively, resulting in text-based vision representation $f^{tbv}$.

\noindent\textbf{TRT and VRT Block.} In these blocks, we transform $f^t$ and $f^v$ into similar structures, thereby fostering a stronger semantic connection between the two modalities. In the TRT block, $f^t$ is multiplied by its transposition ${f^t}^T$ and then transformed through a fully connected (FC) layer and a reshaped layer, yielding $\tilde{f^{t}}$. In the VRT block, $f^v$ is transformed through an FC layer followed by four up-sampling layers to obtain $\tilde{f^{v}}$.



\noindent\textbf{CMAF Block.} Partly inspired by the cross-modal fusion component in the CMAFGAN framework \cite{luo2022cmafgan} which was originally designed for word-to-face synthesis tasks, we identified its potential for modal fusion and enhanced it to suit our task. Additionally, we incorporated a SoftPool layer \cite{stergiou2021refining} into the block to reduce computational overhead while preserving more information. Furthermore, the overall structure of the block was restructured. 

\begin{figure}[!h]
    \begin{center}
    \includegraphics[width=8.6cm]{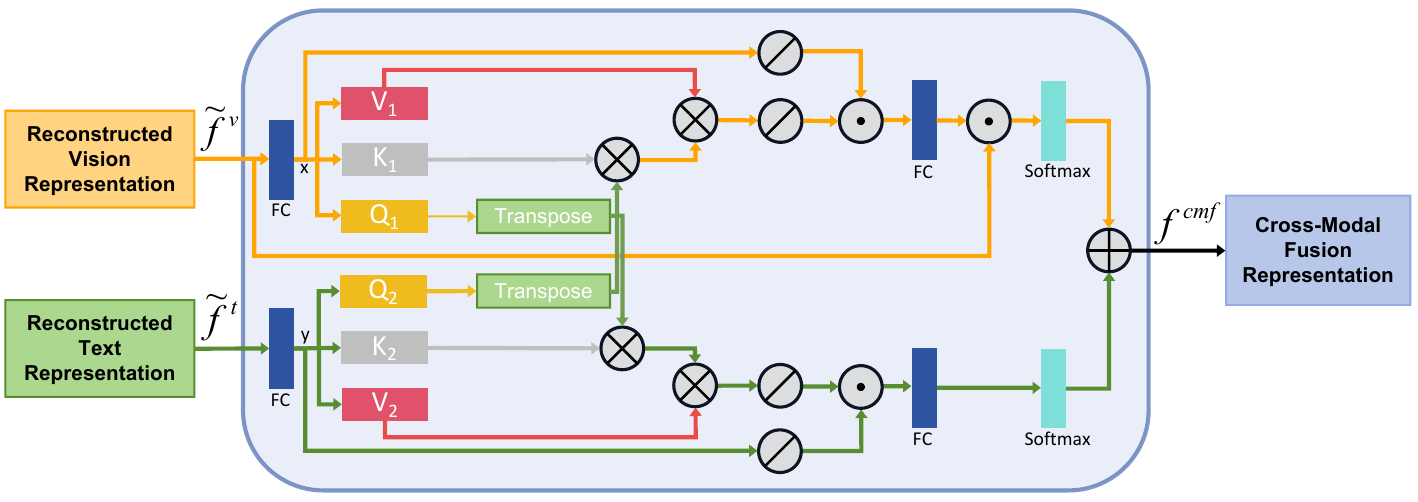}
    \end{center}
   \caption{Architecture of the proposed CMAF Block. $\otimes$ denotes matrix multiplication, $\oslash$ signifies SoftPool, $\odot$ stands for matrix addition, and $\oplus$ is representative of concatenation. }
    \label{CMAF}
\end{figure}

As shown in Fig. \ref{CMAF}, we initially diminish the size of inputs $\tilde{f^v}$ and $\tilde{f^{t}}$ through the FC layer, referring to these as $x$ and $y$. Following this, $x$ and $y$ are separately transformed into three feature spaces via $1\times1$ convolution layers, which are referred to as $V_1$, $K_1$, $Q_1$ and $V_2$, $K_2$, $Q_2$, with $w$ and superscripts denoting their corresponding weight matrices. Then, we can compute the matching degree as follows:
\vspace{-0.1cm}
\begin{eqnarray}
\beta_{j, i}=\frac{\exp \left(\mathbf{s}_{i j}\right)}{\sum_{i=1}^{S} \exp \left(\mathbf{s}_{i j}\right)} \text{, where } \mathbf{s}_{i j}=w^{Q1}x_{i}^{T}\times w^{K2}y_{j},\\
\rho_{j, i}=\frac{\exp \left(\mathbf{t}_{i j}\right)}{\sum_{j=1}^{S} \exp \left(\mathbf{t}_{i j}\right)} \text{, where } \mathbf{t}_{ij}=w^{Q2}y_{i}^{T}\times w^{K1}x_{j},
\end{eqnarray}
where $\beta$ and $\rho$ signify the matching degree in vision and text spaces, separately. We multiplied the matrices $\beta$ and $V_1$,  $\rho$ and $V_2$ to get cross-modal attention feature map $\mathbf{o}_x$ and $\mathbf{o}_y$.

Subsequently, we apply SoftPool to the previously obtained $\mathbf{o}_{x}$, $\mathbf{o}_{y}$, $x$ and $y$ to yield $\Breve{\mathbf{o}}_{x}$, $\Breve{\mathbf{o}}_{y}$, $\Breve{x}$ and $\Breve{y}$. Following this, we add the matrices $\Breve{\mathbf{o}}_{x}$ and $\Breve{x}$, $\Breve{\mathbf{o}}_{y}$ and $\Breve{y}$, and pass them through a linear layer to obtain $\mathbf{o}_{v}$ and $\mathbf{o}_{w}$. Lastly, after applying a softmax layer to each, we can express $f^{cmf}$ as follows:
\begin{eqnarray}
   f^{cmf}=\operatorname{concat}\left(\gamma_{1} * \mathbf{o}_{v}, \gamma_{2} * \mathbf{o}_{w}\right).
\end{eqnarray}

\noindent\textbf{MHSAF Block.} We implemented a multi-head self-attention mechanism to map features to different subspaces via several distinct linear transformations. Subsequently, we execute self-attention computations on each subspace to procure multiple output vectors, which are then concatenated.

\subsection{Loss Function}
In our study, we propose a joint loss function known as the VTMF loss. This loss is composed of three integral components. Firstly, the intra-modality and inter-modality alignment (IMIMA) loss is incorporated as a global loss. Its purpose is to map semantically similar samples from both intra-modalities and inter-modalities into a harmonious global space. Secondly, the similarity distribution matching (SDM) loss is employed to enhance semantic matching and to extract inherent dependencies between the two modalities. Finally, the function includes masked language modeling (MLM) loss, which serves to enrich semantic learning and augment textual comprehension.

\noindent\textbf{IMIMA Loss.} To accomplish alignment on both intra-modalities, such as Text-to-Text ($t2t$) and Vision-to-Vision ($v2v$), as well as inter-modalities, specifically Text-to-Vision ($t2v$) and Vision-to-Text ($v2t$), we map semantically related samples into related individual spaces and maintain the proximity of similar samples in the joint embedding space. We designate the negative sets for the sample as $N$. In intra-modalities is $\mathbf{N}_{i}^{intra} =\left\{y_{j} \mid \forall y_{j} \in N, j \neq i\right\}$ and in inter-modalities is $\mathbf{N}_{i}^{inter} =\left\{x_{j} \mid \forall x_{j} \in N, j \neq i\right\}$. Thus, the intra/inter loss can be expressed as follows:
\begin{eqnarray}
\mathscr{L}^{A2B}_{intra/inter}=-\log \frac{\delta\left(f^A, f^B\right)}{\delta\left(f^A, f^B\right)+ \sum_{f_{k} \in \mathbf{N}} \delta\left(f^A,f_{k}^B\right)},
    \label{imima}
\end{eqnarray}
where $\delta\left(a, b\right)=\exp \left(a^{T} b\right)$. Therefore, IMIMA loss is:
\begin{eqnarray}
\mathscr{L}_{IMIMA}=\mathscr{L}_{intra}^{t2t}+\mathscr{L}_{intra}^{v2v}+\mathscr{L}_{inter}^{t2v}+\mathscr{L}_{inter}^{v2t}.
\end{eqnarray}

\noindent\textbf{SDM Loss.} We employ SDM loss \cite{jiang2023cross} to forge consistent semantic match, thus associating the representations across modalities. For each vision-text pair, we obtain a vision representation $f^v_i$ and a text representation $f^t_j$. And we define $\left\{\left(f_{i}^{v}, f_{j}^{t}\right), l_{i, j}\right\}$, where $l_{i, j}$ is the matching label. When $l_{i,,j}=1$ means that $(f_{i}^{v}, f_{j}^{t})$ is a matched pair which denotes the two models from the same identity, while $l_{i,,j}=0$ indicates the unmatched pair. The true matching probability can be formulated as:
\begin{eqnarray}
    q_{i, j}=l_{i, j} \Biggl/ \sum_{k=1}^{N}l_{i,k}
\end{eqnarray}

Let $\operatorname{sim}(\mathbf{u}, \mathbf{v})=\mathbf{u}^{\top} \mathbf{v} /\|\mathbf{u}\|\|\mathbf{v}\|$ denotes the dot product between $L_2$ normalized $u$ and $v$ (i.e. cosine similarity). The matching probability $p_{i, j}$ can be deemed as the proportion of the cosine similarity score between $f_i^v$ and $f_j^t$ to the sum of the cosine similarity score between $f_i^v$ and $\{f^t_j\}_{j=1}^N$\cite{rahutomo2012semantic}. Then the probability of matching pairs can be simply calculated with the following $softmax$ function \cite{jiang2023cross}:
\begin{eqnarray}
    p_{i, j}=\frac{\exp \left(\operatorname{sim}\left(f_{i}^{v}, f_{j}^{t}\right) / \tau\right)}{\sum_{k=1}^{N} \exp \left(\operatorname{sim}\left(f_{i}^{v}, f_{k}^{t}\right) / \tau\right)}
    \label{pij}
    \end{eqnarray}
where $\tau$ the temperature hyperparameter to limit the probability distribution peaks.

The SDM loss of $v2t$ can be delineated as follows:
\begin{eqnarray}
    \mathscr{L}_{v 2 t}=K L\left(p_i \| q_i \right)=\frac{1}{n} \sum_{i=1}^{n} \sum_{j=1}^{n} p_{i, j} \log \left(\frac{p_{i, j}}{q_{i, j}}\right),
    \label{v2t}
\end{eqnarray}
where $q$ represents the true matching probability and $p$ signifies the proportion of a specific cosine similarity score to the overall sum. The bi-directional SDM loss is the sum of the loss of $v2t$ and $t2v$.


\noindent\textbf{MLM Loss.} We adopt the design of the intrinsic loss function from BERT. The objective of MLM is to randomly obscure certain words in input texts. The model is then required to predict these hidden words, serving for assessment of loss.

\noindent\textbf{Overall Objective.} Based on the analysis above, the definition of VTMF loss can be calculated as follows:
\begin{eqnarray}
\mathscr{L}_{VTMF}=\mathscr{L}_{IMIMA}+\alpha\mathscr{L}_{SDM}+\beta\mathscr{L}_{MLM},
\label{feq}
\end{eqnarray}
where $\alpha$ and $\beta$ represent the weights of $\mathscr{L}_{SDM}$ and $\mathscr{L}_{MLM}$, respectively, serving to dynamically balance the relative significance of these losses.

\section{EXPERIMENT AND RESULTS}
\subsection{Experiment Setting}
\noindent\textbf{Dataset.} In this study, we utilized a private ICH dataset obtained from our collaborative hospital, comprising a total of 294 cases with 149 indicating good and 145 bad prognoses. Each case included comprehensive CT imaging with demographic and clinical information including gender, age, onset-to-CT time, hospital stay, GCS score, and treatment method as well as hemorrhage position and volume. Each case was labeled with either a good or bad prognosis. The classification label is the prognosis outcomes of patients, which is determined by the Glasgow Outcome Scale (GOS) by neurologists. GOS is a rating scale that assesses patients’ functional outcomes following brain injury and then according to it, neurologists can label each sample as good or bad. In terms of data preprocessing,  we carried out several operations, there are the following steps:
(1) Convert series 2D DICOM (Digital Imaging and Communications in Medicine) to 3D NIfTI (Neuroimaging Informatics Technology Initiative) format through dcm2niix. (2) Remove the skull and extract brain tissue with the Swiss Skull Scripper plugin in 3D slicers and the Numpy and Scipy package in Python. (3) Resample the images, constrained HU scales and perform Z-score standardization.

\noindent\textbf{Implementation Details.} Experiments were conducted using two NVIDIA HGX A100 Tensor Core GPUs, employing the Adam optimizer. The training epoch, learning rate, and batch size were respectively set at 300, 0.0001, and 128. Finally, $\alpha$ and $\beta$ in Eq \ref{feq} is learned as 0.84 and 0.45. All experiments were conducted through five-fold cross-validation.

\subsection{Visualization Analysis}
\label{visionanalysis}
To better verify the interpretability of our work, we designed visual experiments. We conducted four comparative experiments, including (a) Good Prognosis Oriented Medical Report, (b) Bad Prognosis Oriented Medical Report, (c) Vision only (the same as the set in Sec.\ref{ablation_test}) and our method. For (a) and (b), we wrote a good prognosis-oriented medical report and a bad prognosis-oriented medical report for the patients by adjusting the patient's demographic and clinical information, with the help of neurosurgeons to simulate the impact of different prognosis-oriented medical reports on the network. The Score-CAM \cite{wang2020score} method is applied to the last convolution layer of Vision Encoder, the last convolution layer of 3D ResNet-50. Especially, because it is a 2D method, we utilized it on the middle and lower 2D slices of visual features (selected the 25th slice out of the 64 slices).

\begin{figure}[]
\center
\includegraphics[width=8cm]{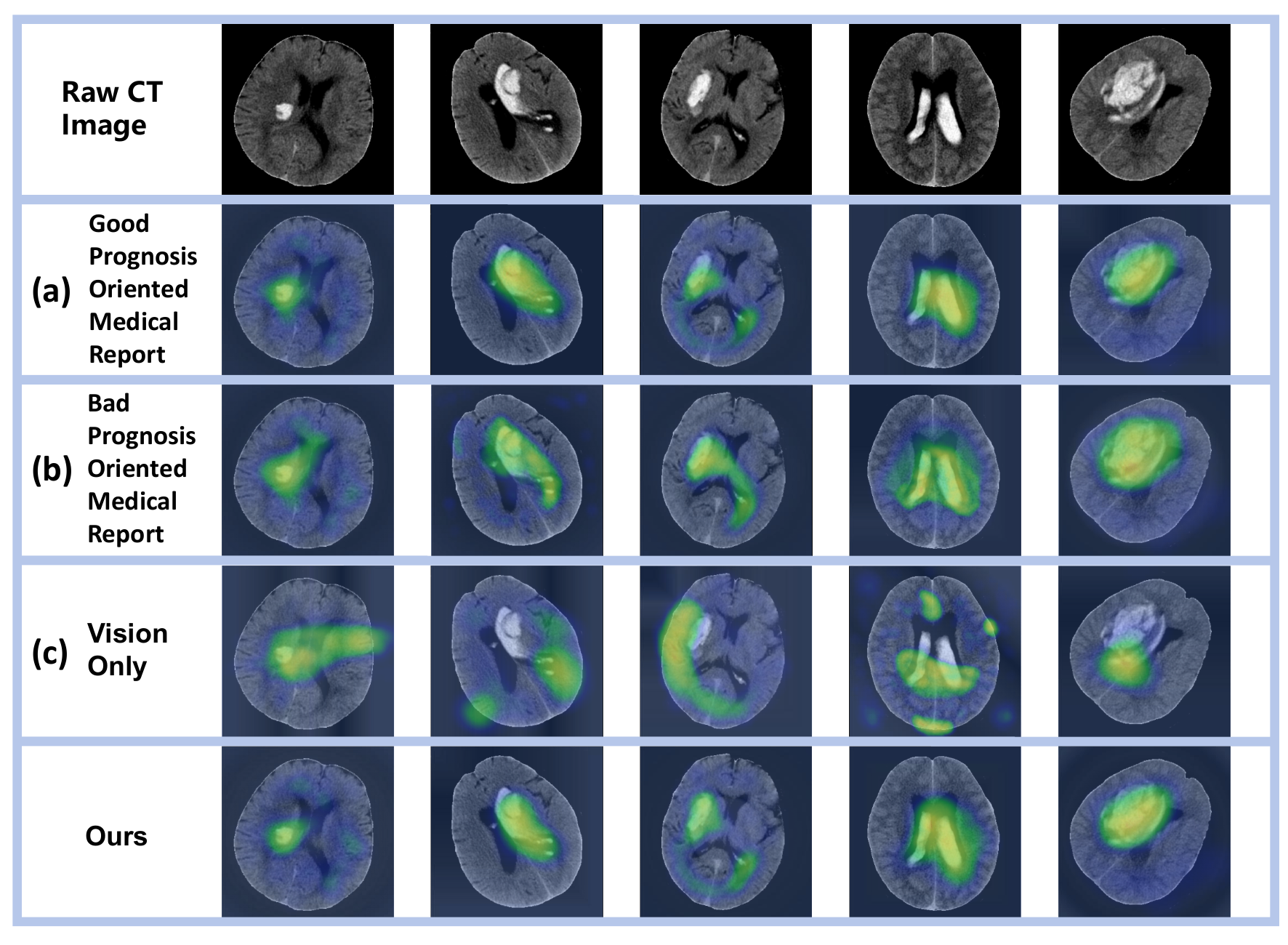}
\caption{This graph depicts the impact of different medical report texts on the regions of interest in the joint-attention mechanism. } 
\label{visual}
\end{figure}

As shown in Fig.\ref{visual}, for different oriented medical reports, our network can accurately locate the region of interest at the location of bleeding, in order to pay more attention to the areas with a higher correlation with prognosis results. For different reports, the region of interest will change to a certain extent with the change of text to match the text information. However, the main part of the region of interest is still determined by the CT image, which is consistent with the original intention of our network design and proves its rationality.
.

\subsection{Ablation Experiment}
\label{ablation_test}
The Text-Only and Vision-Only models directly input the features extracted from their corresponding encoders into the MHSAF block, followed by a classification module. Compared to Vision-Only, ICHPro demonstrates a significant improvement, exceeding it by \textbf{9.79\%} in accuracy and \textbf{0.0834} in AUC metrics. Our method learns modal fusion features that encompass richer demographic and clinical information. This enhances the extraction of more contextual information, thereby facilitating more accurate prognostic predictions.
\begin{table}[H]
\centering
\caption{Results of modal ablation experiment.}
\resizebox{8.6cm}{!}{
\begin{tabular}{cccccc}
\hline
Method & Acc(\% )& Recall(\%) & Prec(\%)  & $F1$ Score & AUC \\\hline 
Text-Only & 69.15 & 65.10 & 71.11 & 0.6797 & 0.7534     \\
Vision-Only & \underline{79.32} & \underline{77.18} & \underline{82.72} & \underline{0.7985} & \underline{0.8595}  \\ \hline
\textbf{ICHPro} & \textbf{89.11} & \textbf{84.56} & \textbf{91.02} & \textbf{0.8767} & \textbf{0.9429} \\
\hline
\end{tabular}}
\label{modalabaltion}
\vspace{-1em}
\end{table}

\subsection{Attention Fusion Structure Experiment}
We further conducted a comparative analysis of six methods, each comprising different permutations and combinations of CMAF and MHSAF blocks. We utilized the terms "Cross" and "Self" to individually denote these blocks. It is important to note that, the notation A-B implies that Block A is entered first, followed by Block B.

\begin{table}[H]
\centering
\caption{Comparisons of attention fusion methods.}
\resizebox{8.6cm}{!}{
\begin{tabular}{cccccc}
\hline
 Structure & Acc(\% )& Recall(\%) & Prec(\%)  & $F1$ Score & AUC  \\\hline
Self Attention & 82.71 & 78.52 & 83.17 & 0.8078 & 0.8671 \\ 
Cross Attention & 84.41 & 79.19 & 86.32 & 0.8260 & 0.8956 \\
Self-Self Attention & 76.94 & 71.81 & 80.95 & 0.7611 & 0.8025\\
Cross-Cross Attention& \underline{87.11} & 80.53 & \underline{89.40} & \underline{0.8473} & \underline{0.9108} \\
Self-Cross Attention & 85.08 & \underline{81.21} & 88.08 & 0.8451 & 0.8934    \\ \hline
\textbf{Cross-Self Attention} & \textbf{89.11} & \textbf{84.56} & \textbf{91.02} & \textbf{0.8767} & \textbf{0.9429}    \\ \hline
\end{tabular}}
\label{attentionmethod}
\vspace{-1em}
\end{table}

Methods incorporating cross-modal attention demonstrate superior performance compared to those lacking this addition, thereby confirming the effectiveness of the CMAF block. As indicated in Table \ref{attentionmethod}, sequentially passing through the CMAF and MHSAF blocks yields optimal results. The former facilitates interaction between two modalities, establishing semantic connections and enriching feature expressions, while the latter captures the internal dependencies of fused features, thereby effectively capturing contextual relationships. This combination significantly amplifies the expressive power and generalization capabilities of cross-modal networks.

\vspace{-0.25em}
\subsection{Loss Function Based Experiment}
We employed three alternative loss functions for the comparative analysis of our model. These included two single cross-modal losses, $\mathscr{L}_{blend}$ and $\mathscr{L}_{cmpm}$, and one joint cross-modal loss, $\mathscr{L}_{CMFA}$. Additionally, we conducted ablation experiments to demonstrate the effectiveness of each component.

\begin{table}[!h]
\centering
\caption{Results of comparison and ablation experiment based on loss function.}
\resizebox{8.6cm}{!}{
\begin{tabular}{ccccccccc}
\hline
& \multicolumn{3}{c}{Components}&&&&& \\ \cline{2-4}
\multirow{-2}{*}{{Loss Function}} & IMIMA & SDM & MLM & \multicolumn{1}{c}{\multirow{-2}{*}{Acc(\%)}} & \multicolumn{1}{c}{\multirow{-2}{*}{Recall(\%)}} & \multicolumn{1}{c}{\multirow{-2}{*}{Prec(\%)}} & \multicolumn{1}{c}{\multirow{-2}{*}{$F1$ Score}} & \multirow{-2}{*}{{AUC}} \\ \hline
$\mathscr{L}_{blend}$\cite{wang2020makes}&&&&74.57&70.47&78.17&0.7412&0.7598\\
$\mathscr{L}_{cmpm}$\cite{zhang2018deep}&&&&73.56&69.13&76.78&0.7275&0.7852\\\cline{1-9} $\mathscr{L}_{CMFA}$\cite{farooq2022axm}&&&&\underline{85.08}&80.54&\underline{88.72}&\underline{0.8442}&\underline{0.8930}\\
\cline{1-9}
$\mathscr{L}_{IMIMA}$&$\surd$&&&75.59&71.14&79.44&0.7506&0.7982\\
$\mathscr{L}_{SDM}$&&$\surd$&&71.86&68.45&73.19&0.7074&0.7346\\
$\mathscr{L}_{MLM}$&&&$\surd$&54.24&50.34&56.94&0.5344&0.5705\\
$\mathscr{L}_{IMIMA}+\alpha\mathscr{L}_{SDM}$&$\surd$&$\surd$&&84.40& \underline{81.88} &86.49&0.8412&0.8806\\
$\mathscr{L}_{IMIMA}+\beta\mathscr{L}_{MLM}$&$\surd$&&$\surd$&76.27&73.83&79.02&0.7634&0.8194\\
 \hline
$\bm{\mathscr{L}_{VTMF}}$& $\surd$&$\surd$&$\surd$ & \textbf{89.11} & \textbf{84.56} & \textbf{91.02} & \textbf{0.8767} & \textbf{0.9429}\\ \hline
\end{tabular}}
\label{loss}
\vspace{-0.25em}
\end{table}

As summarized in Table \ref{loss}, joint losses, which amalgamate multiple optimization objectives, yield superior performance compared to single losses. As our global loss, $\mathscr{L}_{IMIMA}$ outperforms the other four single losses due to its ability to align both intra and inter-modal. Although the individual $\mathscr{L}_{MLM}$ performs poorly when paired with losses bearing cross-modal capabilities, it can effectively enhance the contextual understanding of fused features. Compared to using $\mathscr{L}_{IMIMA}$ independently, the addition of $\mathscr{L}_{SDM}$ or $\mathscr{L}_{MLM}$ increases accuracy by 8.81\% and 0.68\%, respectively. This demonstrates the efficacy of our additions. These findings suggest that the combination of all three losses can achieve optimal performance for each individual loss. Compared to another joint loss function $\mathscr{L}_{CMFA}$,  $\mathscr{L}_{VTMF}$ outperforms by 4.03\% and 0.0499 in accuracy and AUC value, respectively. This analysis highlights the effectiveness of our loss function.

\vspace{-0.25em}
\subsection{Comparative Experiments}

\begin{figure}[h]
\center
\includegraphics[width=0.304\textwidth]{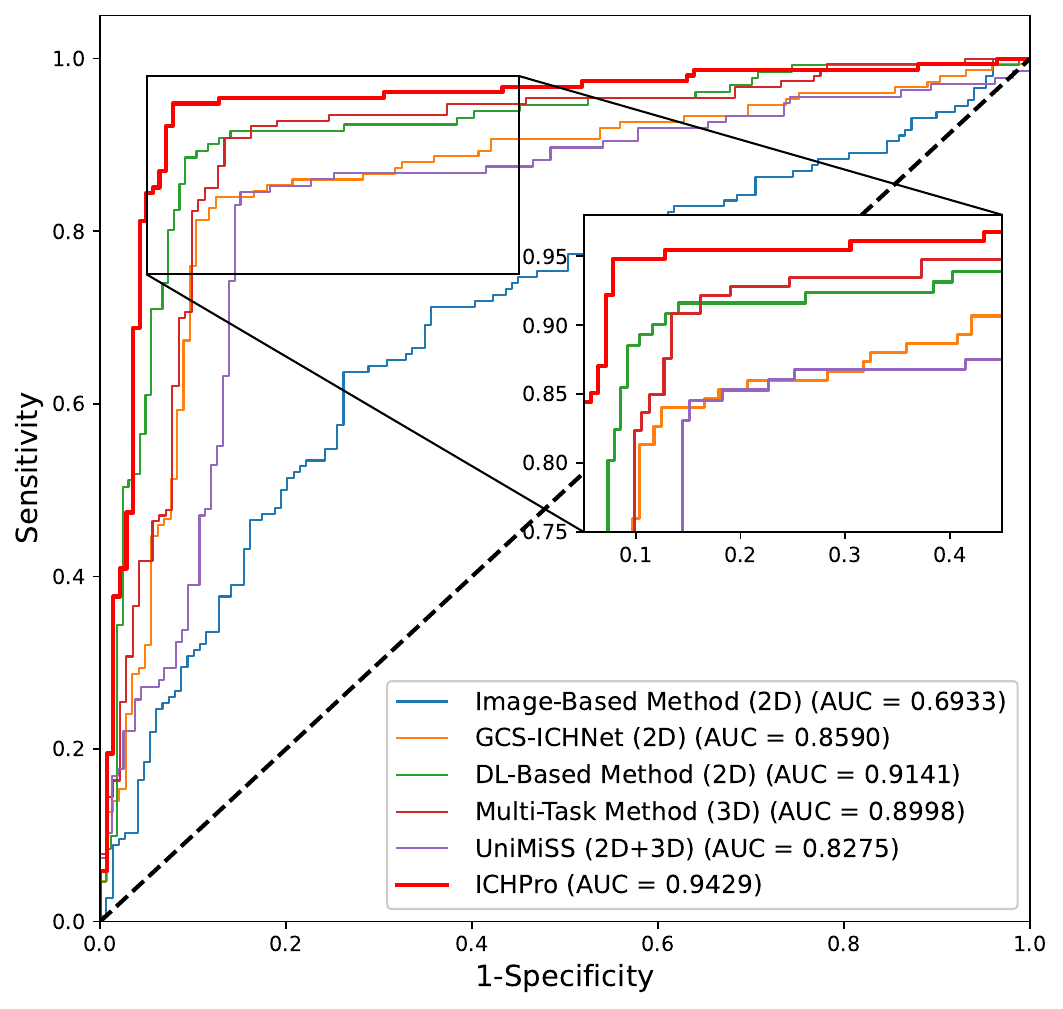}
\caption{ROC curves between ICHPro and other methods.} \label{netsroc}
\end{figure}

We conducted a comparison of ICHPro with other advanced methods, using our dataset. The results are illustrated in Table \ref{comparision} and Fig.\ref{netsroc}. The first four methods delineated in the table are specifically designed for the classification of ICH prognosis, while UniMiSS represents a universal network for medical image classification that utilizes a combination of 2D and 3D convolutional techniques.


\begin{table}[]
\centering
\caption{Comparisons of ICHPro and other methods.}
\resizebox{8.6cm}{!}{
\begin{tabular}{cccccc}
\hline
Method & Acc(\%) & Recall(\%) & Prec(\%) & $F1$ Score & AUC \\\hline
Image-Based Method (2D) \cite{nawabi2021imaging} & 74.23 & 67.11 & 75.98 & 0.7127 & 0.6933\\
GCS-ICHNet (2D) \cite{shan2023gcs} & 85.08 & \underline{81.88} & 87.25 & 0.8448 & 0.8590\\
DL-Based Method (3D) \cite{perez2023deep} & 81.02 & 78.52 & 83.31 & 0.8084 &  \underline{0.9141} \\
Multi-Task Method (3D) \cite{gong2023unified} & \underline{85.42} & 79.86 & \underline{89.80} & \underline{0.8454} & 0.8998 \\
UniMiSS (2D+3D) \cite{xie2022unimiss} & 82.03 & 78.52 & 87.59 & 0.8281 & 0.8275 \\
\hline 
\textbf{ICHPro} & \textbf{89.11} & \textbf{84.56} & \textbf{91.02} & \textbf{0.8767} & \textbf{0.9429}\\ \hline
\end{tabular}}
\label{comparision}
\vspace{-1.5em}
\end{table}

Owing to the integration of domain knowledge, the accuracy of the 2D GCS-ICHNet essentially matches that of the 3D multi-task method, which relies solely on images. It also surpasses the universally applied 2D+3D UniMiSS method. Both ICHPro and the DL-based method incorporate comprehensive demographic and clinical information, rendering their AUC superior to all other methods. This highlights the enhanced robustness of networks that fuse information beyond images. Compared to the DL-based method, our performance is superior across all metrics, underscoring the effectiveness of our joint-attention fusion mechanism. When compared to the optimal indicators of other methods, ours improves accuracy by \textbf{3.69\%} and the AUC value by \textbf{0.0288}. In addition, the ROC curve is closest to the upper left corner, indicative of its effectiveness in distinguishing between positive and negative samples. Our method demonstrates a comprehensive superiority over the comparison methods, and to our knowledge, it surpasses existing advanced methods in the task of ICH prognosis classification on our dataset.

\section{CONCLUSION}
The absence of demographic and clinical information and the inefficient cross-modal fusion mechanism could hinder the effective extraction of cross-modal fusion features. To address this, in this paper, we proposed an ICHPro, a joint-attention fusion-based 3D cross-modal network, for ICH prognosis classification. Furthermore, we proposed a VTMF loss to enhance modal alignment and optimize networks. Our experimental results demonstrate the efficacy of our method. In the future, we aim to extend the network to an end-to-end model and augment classification tasks through segmentation, for improved outcomes. Additionally, our proposed method holds potential for application beyond ICH prognosis, extending to other medical cross-modal classification tasks.


\section{COMPLIANCE WITH ETHICAL STANDARDS}
\begin{sloppypar}
This study was conducted by the principles of the Declaration of Helsinki. Approval was granted by the Ethics Committee of Longgang Central Hospital of Shenzhen (2023.10.26/No.2023ECPJ077).
\end{sloppypar}
\vspace{-0.3em}
\section{ACKNOWLEDGMENTS}
Support of the Zhejiang Provincial Natural Science Foundation of China (No.LY21F020017,2022C03043), Joint Funds of the Zhejiang Provincial Natural Science Foundation of China (No.U20A20386), National Natural Science Foundation of China (No.61702146), GuangDong Basic and Applied Basic Research Foundation (No.2022A1515110570) and Innovation Teams of Youth Innovation in Science and Technology of High Education Institutions of Shandong Province (No.2021KJ088) are gratefully acknowledged.

\vspace{-0.3em}
\bibliographystyle{IEEEbib}
\bibliography{refs}

\end{document}